\documentclass[runningheads]{llncs}
\usepackage[T1]{fontenc}
\usepackage{graphicx}
\usepackage{hyperref}
\usepackage[table]{xcolor}
\usepackage{caption}
\usepackage{color}
\usepackage{makecell}
\usepackage{amsfonts}
\usepackage{amsmath}
\usepackage{float}
\usepackage{diagbox}
\usepackage{tabularx}
\usepackage{multirow}

\usepackage{booktabs} 
\usepackage{graphicx}
\usepackage{subcaption}
\usepackage{bbm}
\usepackage{diagbox}  

\begin{document}
\title{Towards Practical Algorithm Selection for Unsupervised Domain Adaptation in Medical Imaging}
\titlerunning{Towards Practical Algorithm Selection for UDA in Medical Imaging}
%
\author{Yiheng Xiong\inst{1} \and
Luisa Gallée\inst{1} \and
Daniel Santak Wolf\inst{1,2} \and
Heiko Hillenhagen\inst{1} \and
Michael Götz\inst{1}
}
\authorrunning{Y. Xiong et al.}
%
\institute{Section of Experimental Radiology, Ulm University Medical Center, Germany \and
Visual Computing Group, Ulm University, Germany\\
\email{yiheng.xiong@uni-ulm.de}}
\maketitle              
\begin{abstract}
Numerous unsupervised domain adaptation (UDA) algori-thms exist, but for clinical practice, selecting the best-suited one along with proper hyperparameters often remains unclear, as the unlabeled deployment (target) domain prevents direct evaluation. We propose a label-free criterion that jointly selects the algorithm and hyperparameters for UDA. Given a pool of candidate models from multiple algorithms trained with different hyperparameters, our approach scores each candidate against an agreement reference, and selects the one with the highest score. The agreement reference is constructed in two levels without using target labels. First, we leverage multiple label-free selection signals, using each to nominate a model within every algorithm. Second, the nominated models are aggregated across algorithms to form a reference prediction for each unlabeled target sample. The candidate whose predictions agree most with this reference is then selected for deployment. Experimental results on four brain MRI and four chest X-ray datasets across seven clinically relevant transfer scenarios show that our method achieves better selection performance than individual validators and remains effective across different algorithm pools. Our approach takes a step towards practical, label-free algorithm selection for clinical deployment of UDA. Our code is open sourced at \href{https://github.com/xiongyiheng/Complete-Unsupervised-Domain-Adaptation-Pipeline/tree/main/two-level-algorithm-selection}{Two-Level-Agreement}.

\end{abstract}
\section{Introduction}
Unsupervised domain adaptation (UDA) aims to transfer knowledge from labeled source data to unlabeled target data, with recent algorithms showing robust adaptation performance on medical imaging benchmarks~\cite{kumari2024deep,xiong2026how}. However, in clinical practice, selecting the best-suited algorithm often remains unclear, since the absence of target labels prevents direct evaluation. Existing unsupervised selection approaches largely consist of validators~\cite{musgrave2022three,saito2021tune,you2019towards,sugiyama2007covariate,morerio2017minimal,tuassessing,hu2023mixed,yang2024can}, label-free proxies that score each candidate checkpoint (a UDA model snapshot from a training iteration) and select the best-scoring one, achieving reliable selection on general-vision UDA benchmarks such as VisDA~\cite{peng2017visda}. However, two aspects remain less explored. First, each checkpoint is typically scored independently by a single validator, without exploiting the valuable agreement among different validators or among different candidate checkpoints. Hu et al.~\cite{hu2024towards} combine predictions from validator-selected checkpoints within individual algorithms to leverage multiple validator signals, but do not exploit agreement across different algorithms. Second, while these methods are effective for hyperparameter selection given a fixed algorithm, the more realistic problem of joint algorithm and hyperparameter selection has received much less attention. Yang et al.~\cite{yang2024can} address algorithm selection by first choosing an algorithm under fixed hyperparameters and then selecting hyperparameters within it, but the joint selection of both, as encountered in clinical practice, remains largely unexplored.

To address this, given a pool of candidate checkpoints from multiple UDA algorithms trained with different hyperparameters, we propose a label-free selection criterion that scores each candidate against an agreement reference and selects the one with the highest score. The agreement reference is constructed in two levels. At the first level, we leverage multiple existing validators, using each to nominate a checkpoint within each algorithm; at the second level, the nominated checkpoints are aggregated across algorithms to form a reference prediction for each unlabeled target sample. The candidate whose predictions agree most with this reference is then selected for deployment.

We validate our method by selecting from multiple UDA algorithms across seven clinically relevant transfer scenarios on four brain MRI and four chest X-ray (CXR) datasets. Experimental results show that our method achieves better selection results than individual validators on average and in six of the seven scenarios. Compared to the best individual validator, it also reduces the target performance gap to the best available checkpoint by more than a third. Our approach also remains effective under different algorithm pools and checkpoint densities. We summarize our contributions as:

\begin{itemize}
\item We propose a label-free criterion for joint algorithm and hyperparameter selection in UDA for medical imaging, which to our knowledge is the first to address this joint problem. Each candidate checkpoint is scored against an agreement reference to identify a single high-performing one across different UDA algorithms, supporting more reliable algorithm selection for clinical deployment where target labels are unavailable.

\item We design a two-level construction of this agreement reference, in which validators select checkpoints within each algorithm at the first level, and their agreement across algorithms forms the reference at the second level, yielding a robust selection signal across diverse algorithm pools.
\end{itemize}

\section{Methodology}

\subsection{Preliminary}

\noindent\textbf{UDA \& Algorithm Selection in UDA.} In UDA, a labeled source domain $\mathcal{D}_s=\{(x_i^s,y_i^s)\}$ and an unlabeled target domain $\mathcal{D}_t=\{x_j^t\}$ are given, and the goal is to train a model using both $\mathcal{D}_s$ and $\mathcal{D}_t$ to make predictions on $\mathcal{D}_t$. We focus on UDA for classification, where a UDA algorithm is trained by jointly minimizing a supervised classification loss $\mathcal{L}_{\text{cls}}$ on the source domain and an adaptation loss $\mathcal{L}_{\text{adapt}}$ that aligns the two domains,
\begin{equation}
\mathcal{L} = \mathcal{L}_{\text{cls}} + \lambda\,\mathcal{L}_{\text{adapt}},
\end{equation}
where $\lambda$ controls the adaptation strength. 

In practice, many UDA algorithms exist, each with different $\mathcal{L}_{\text{adapt}}$ and typically trained under several hyperparameter configurations (such as different values of $\lambda$), with model snapshots saved at multiple iterations during training. We refer to each such snapshot as a checkpoint, so that a checkpoint is uniquely identified by an algorithm, a hyperparameter configuration, and a training iteration. Let $\mathcal{A}=\{a_1,\dots,a_M\}$ denote a set of UDA algorithms, and let $\Theta$ denote the full pool of resulting checkpoints across all algorithms, hyperparameter configurations, and training iterations. Since target labels are unavailable, the target performance of any checkpoint $\theta\in\Theta$ cannot be measured directly, and it is therefore unknown which checkpoint performs best on $\mathcal{D}_t$. The task of algorithm selection is to identify a single checkpoint $\hat{\theta}\in\Theta$ for deployment, where the candidates span different UDA algorithms, hyperparameter configurations, and training iterations rather than being restricted to a single algorithm.

\noindent\textbf{Validators.} A validator is a function $v$ that maps a checkpoint $\theta$ to a scalar score $v(\theta)\in\mathbb{R}$, computed without target labels, and serves as a proxy for target performance, so that the checkpoint with the best score is selected. The set of validators $\mathcal{V}=\{v_1,\dots,v_N\}$ can roughly be divided into two categories. Source-guided includes Source-Risk~\cite{ganin2015unsupervised}, IWCV~\cite{sugiyama2007covariate}, DEV~\cite{you2019towards}, and DEV-N~\cite{musgrave2022three}. Target-specific includes Entropy~\cite{morerio2017minimal}, InfoMax~\cite{musgrave2021unsupervised}, MCC (V)~\cite{jin2020minimum}, BNM (V)~\cite{musgrave2022three}, Corr-C~\cite{tuassessing}, SND~\cite{saito2021tune}, Class-AMI~\cite{musgrave2022three}, MixVal~\cite{hu2023mixed} and TransScore~\cite{yang2024can}.

\subsection{Two-Level Agreement Reference Construction}

\begin{figure*}
    \centering
    \includegraphics[width=1.0\linewidth]{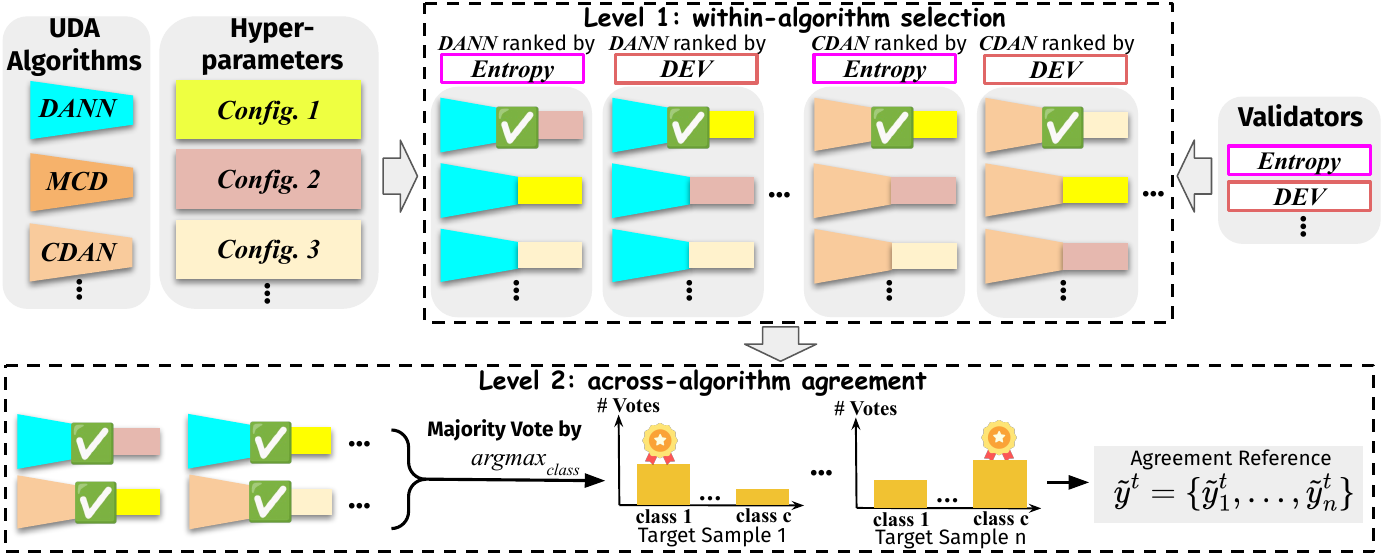}
    \caption{Two-level agreement reference construction. Each candidate combines an algorithm (trapezoid) and a hyperparameter configuration (rectangle). At Level 1, each validator nominates the top checkpoint within each algorithm (green check). At Level 2, the predicted class probabilities of the nominated checkpoints are averaged per target sample, and the class with the highest average (medal) forms the agreement reference $\tilde{y}^t$.}
    \label{fig:agreement_construction}
\end{figure*}

Since target labels are unavailable, no direct reference exists for comparing checkpoints. We therefore construct a reference from the candidates themselves, in two levels, as illustrated in Figure~\ref{fig:agreement_construction}. Each checkpoint $\theta\in\Theta$ produces a prediction $\hat{y}^t_j(\theta)=\arg\max_c p_c(x_j^t;\theta)$ for every target sample $x_j^t$, where $p_c(\cdot;\theta)$ denotes the predicted probability of class $c$.

\noindent\textbf{Level 1: within-algorithm selection.} For each algorithm $a\in\mathcal{A}$, every validator $v\in\mathcal{V}$ scores the checkpoints of $a$ and nominates the best one,
\begin{equation}
\theta_{a,v}=\arg\max_{\theta\in\Theta_a} v(\theta),
\end{equation}
where $\Theta_a=\{\theta_{a,k}\mid k=1,\dots,K\}$ denotes the checkpoints of algorithm $a$. This yields a set of validator-nominated checkpoints $\{\theta_{a,v}\mid v\in\mathcal{V}\}$ for each algorithm, where each checkpoint reflects the unique selection signal of a validator. 

\noindent\textbf{Level 2: across-algorithm agreement.} The validator-nominated checkpoints across algorithms, $\{\theta_{a,v}\mid a\in\mathcal{A},\,v\in\mathcal{V}\}$, are then combined by averaging their predicted class probabilities. For each target sample $x_j^t$, the agreement reference is
\begin{equation}
\tilde{y}^t_j=\arg\max_{c}\frac{1}{|\mathcal{A}||\mathcal{V}|}\sum_{a\in\mathcal{A}}\sum_{v\in\mathcal{V}}p_c(x_j^t;\theta_{a,v}),
\end{equation}
where a checkpoint nominated by several validators contributes once per nomination. By selecting within each algorithm using each validator and then aggregating across algorithms, the reference draws on the complementary strengths of different UDA algorithms and validators, rather than relying on any single one.

\subsection{Label-Free Selection Criterion}

Given the agreement reference $\tilde{y}^t=\{\tilde{y}^t_j\}_{j=1}^{|\mathcal{D}_t|}$ as a label-free target, every candidate checkpoint in the pool $\Theta$ is scored by how closely its predictions match the reference $\tilde{y}^t$. The agreement score of a checkpoint $\theta$ is measured as the per-class fraction of target samples on which its prediction coincides with the reference, averaged over all classes,
\begin{equation}
s(\theta)=\frac{1}{C}\sum_{c=1}^{C}\frac{\sum_{j}\mathbbm{1}\!\left[\hat{y}^t_j(\theta)=c\right]\mathbbm{1}\!\left[\tilde{y}^t_j=c\right]}{\sum_{j}\mathbbm{1}\!\left[\tilde{y}^t_j=c\right]},
\end{equation}
where $\mathbbm{1}[\cdot]$ is the indicator function and $C$ is the number of classes. The single checkpoint with the highest agreement score is selected for deployment,
\begin{equation}
\hat{\theta}=\arg\max_{\theta\in\Theta}\,s(\theta).
\end{equation}
The selected $\hat{\theta}$ is a single checkpoint from one UDA algorithm, obtained without target labels. The agreement reference is used only to identify it, without the need to deploy all the nominated checkpoints.

\section{Experiments}

\noindent\textbf{Datasets.} To construct medical UDA scenarios, four widely adopted brain MRI datasets are used: ADNI-1, ADNI-2, ADNI-3~\cite{jack2008alzheimer}, and AIBL~\cite{ellis2009australian}. In addition, four publicly available CXR datasets are used: RSNA~\cite{wang2017chestx}, Child CXR~\cite{kermany2018identifying}, LDD~\cite{ldd}, and CRD~\cite{crd}. Each dataset is treated as a separate domain, and transfer is performed within each modality. The brain MRI datasets contain Alzheimer's disease and cognitively normal subjects, while the CXR datasets contain pneumonia and non-pneumonia subjects. Both modalities undergo standard preprocessing following prior work~\cite{guan2021multi,zhang2020collaborative}. Dataset statistics are summarized in Table~\ref{tab:data_statistics}.

\begin{table}[h]
\centering
\caption{Dataset statistics for brain MRI and CXR datasets. Values are the number of samples per class. AD: Alzheimer's disease; CN: cognitively normal; Pneu.: pneumonia; Non-Pneu.: non-pneumonia.}
\label{tab:data_statistics}
\setlength{\tabcolsep}{2pt}
\renewcommand{\arraystretch}{1.0}
\scriptsize
\begin{tabular}{l l l l}
\toprule
\multicolumn{2}{c}{\textbf{Brain MRI}} & \multicolumn{2}{c}{\textbf{CXR}} \\
\cmidrule(lr){1-2} \cmidrule(lr){3-4}
\textbf{Dataset} & \textbf{AD / CN} & \textbf{Dataset} & \textbf{Pneu. / Non-Pneu.} \\
\midrule
ADNI-1~\cite{jack2008alzheimer} & 200 / 221 & RSNA~\cite{wang2017chestx} & 6,012 / 20,672 \\
ADNI-2~\cite{jack2008alzheimer} & 159 / 232 & Child CXR~\cite{kermany2018identifying} & 4,273 / 1,583 \\
ADNI-3~\cite{jack2008alzheimer} & 85 / 431 & LDD~\cite{ldd} & 5,776 / 3,919 \\
AIBL~\cite{ellis2009australian} & 78 / 477 & CRD~\cite{crd} & 9,237 / 10,319 \\
\bottomrule
\end{tabular}
\end{table}

\noindent \textbf{UDA Algorithms.} Algorithm selection is performed over a pool of ten established UDA algorithms spanning multiple paradigms: feature-distance minimization (MMD~\cite{long2015learning}), adversarial alignment (DANN~\cite{ganin2016domain}, CDAN~\cite{long2018conditional}, and DALN~\cite{chen2022reusing}), information maximization (MCC~\cite{jin2020minimum}), SVD loss (BNM~\cite{cui2020towards}), pseudo labeling (ATDOC~\cite{liang2021domain}), classifier discrepancy (MCD~\cite{saito2018maximum}), and medical-specific AD2A~\cite{guan2021multi} (brain MRI) and CoUDA~\cite{zhang2020collaborative} (CXR).

\noindent\textbf{Experimental Setup.} Following Guan et al.~\cite{guan2021multi}, five UDA scenarios (sourc-e$\rightarrow$target) are constructed for brain MRI: ADNI-1$\rightarrow$ADNI-2, ADNI-1$\rightarrow$ADNI-3, ADNI-2$\rightarrow$ADNI-1, ADNI-2$\rightarrow$ADNI-3, and ADNI-1+2$\rightarrow$AIBL. Following Feng et al.~\cite{feng2023contrastive} and Liu et al.~\cite{liu2023attention}, two scenarios are constructed for CXR: RSNA$\rightarrow$Child CXR and LDD$\rightarrow$CRD. Following Musgrave et al.~\cite{musgrave2021unsupervised}, both source and target data are split into training and validation sets, and target performance is measured on the target validation set using balanced accuracy. Stratified five-fold cross-validation is performed, and the mean $\pm$ standard deviation is reported. We do not hold out a separate target test set, as the selection procedure is label-free: given any new unlabeled target data, the validation scores can be recomputed directly on it, and the best-scored checkpoint selected.

\noindent\textbf{Implementation Details.} For all experiments, the classification head is a three-layer MLP with a dropout rate of 0.5. For brain MRI, a 3D ResNet-50~\cite{he2016deep} trained from scratch is used as the backbone, with a batch size of eight per domain. For CXR, a 2D ResNet-50 pretrained on ImageNet~\cite{deng2009imagenet} is used for RSNA$\rightarrow$Child CXR and a pretrained 2D DenseNet-121~\cite{huang2017densely} for LDD$\rightarrow$CRD, both with a batch size of 48 per domain. The two backbones let us assess whether our selection criterion remains robust under a different architecture. All algorithms are trained with AdamW (weight decay 1e-4) and a one-cycle learning rate schedule with warm-up and a peak learning rate of 1e-3, for 10k iterations on brain MRI and 30k iterations on CXR. The adaptation strength is varied over $\lambda\in\{0.1, 0.5, 1.0\}$, giving three runs per algorithm. After warm-up, checkpoints are saved at uniform intervals, yielding 50 checkpoints per run and 150 checkpoints per algorithm. $\mathcal{L}_{\text{cls}}$ is cross-entropy with class-balanced weighting derived from the source labels. All runs are conducted on an A6000 GPU. Training an algorithm for one fold and one value of $\lambda$ takes roughly 6 to 15 hours on brain MRI and 2 to 5 hours on CXR, depending on the algorithm. The output logits of each checkpoint are saved during training, so running the validators requires no additional forward passes.

\section{Results and Discussion}

\noindent\textbf{Main Results.} Table~\ref{tab:main_results} reports the selection results on all UDA scenarios. Due to space constraints, we show the top five individual validators ranked by overall average performance. Our method achieves the highest overall average accuracy (84.5\%), improving over the best individual validator (DEV-N, 81.5\%) by 3.0\%, and reduces the gap to the \textit{Oracle}, the best checkpoint identifiable with target labels, from 9.3\% to 6.4\%. The improvement holds across most individual scenarios: our method ranks first on three of the five brain MRI scenarios and on both CXR scenarios, while DEV-N is better on ADNI-2$\rightarrow$ADNI-1 (82.8\% vs.\ 82.5\%) and DEV on ADNI-1+2$\rightarrow$AIBL (89.0\% vs.\ 86.8\%). On CXR, where the two scenarios use different backbones, our method also obtains the best average, indicating that the selection criterion remains effective under a different architecture.

\begin{table}[h]
\centering
\caption{Selection results across UDA scenarios on brain MRI and CXR, reported as target accuracy (\%). \textit{Oracle} selects the best checkpoint using target labels and serves as an upper bound. \textbf{Bold} indicates the best non-\textit{Oracle} result. Our method achieves the highest average accuracy and the smallest gap to \textit{Oracle}.}
\label{tab:main_results}
\setlength{\tabcolsep}{1pt}
\renewcommand{\arraystretch}{1.0}
\footnotesize
\resizebox{\textwidth}{!}{%
\begin{tabular}{l | cccccc | c ccc| c cc}
\toprule
& \multicolumn{6}{c}{\textbf{Brain MRI Scenarios}} & & \multicolumn{3}{c}{\textbf{CXR Scenarios}} & & \multicolumn{2}{c}{\textbf{Overall}} \\
\cmidrule(lr){2-7} \cmidrule(lr){9-11} \cmidrule(lr){13-14}
Validator
& A1$\rightarrow$A2 & A1$\rightarrow$A3 & A2$\rightarrow$A1 & A2$\rightarrow$A3 & A1+2$\rightarrow$AIBL & \textbf{Avg.}
& & RSNA$\rightarrow$Child & LDD$\rightarrow$CRD & \textbf{Avg.}
& & All & $\Delta$\textit{Oracle}$\downarrow$ \\
\midrule
\rowcolor{gray!20}
\textit{Oracle} & 93.2\,\scriptsize{$\pm$2.0} & 94.2\,\scriptsize{$\pm$3.1} & 89.1\,\scriptsize{$\pm$3.4} & 92.0\,\scriptsize{$\pm$2.8} & 93.8\,\scriptsize{$\pm$2.1} & 92.4 & & 89.7\,\scriptsize{$\pm$0.40} & 83.8\,\scriptsize{$\pm$0.69} & 86.7 & & 90.8 & 0.0 \\
\midrule
Class-AMI & 81.0\,\scriptsize{$\pm$2.2} & 80.6\,\scriptsize{$\pm$3.6} & 79.4\,\scriptsize{$\pm$4.1} & 81.6\,\scriptsize{$\pm$6.3} & 83.2\,\scriptsize{$\pm$3.6} & 81.2 & & 74.2\,\scriptsize{$\pm$8.7} & 78.0\,\scriptsize{$\pm$1.8} & 76.1 & & 79.7 & 11.1 \\
InfoMax & 83.2\,\scriptsize{$\pm$7.2} & 70.6\,\scriptsize{$\pm$7.9} & 82.5\,\scriptsize{$\pm$4.9} & 81.4\,\scriptsize{$\pm$3.1} & 81.6\,\scriptsize{$\pm$5.0} & 79.9 & & 78.8\,\scriptsize{$\pm$5.9} & 80.5\,\scriptsize{$\pm$1.7} & 79.6 & & 79.8 & 11.0 \\
Source-Risk & 84.4\,\scriptsize{$\pm$2.5} & 83.6\,\scriptsize{$\pm$4.8} & 78.9\,\scriptsize{$\pm$4.6} & 81.7\,\scriptsize{$\pm$8.8} & 84.0\,\scriptsize{$\pm$5.1} & 82.5 & & 75.3\,\scriptsize{$\pm$4.1} & 77.3\,\scriptsize{$\pm$1.6} & 76.3 & & 80.7 & 10.1 \\
DEV & 84.4\,\scriptsize{$\pm$5.9} & 82.3\,\scriptsize{$\pm$7.1} & 79.9\,\scriptsize{$\pm$2.1} & 82.3\,\scriptsize{$\pm$7.7} & \textbf{89.0}\,\scriptsize{$\pm$1.7} & 83.6 & & 72.6\,\scriptsize{$\pm$12} & 75.5\,\scriptsize{$\pm$7.9} & 74.1 & & 80.9 & 10.0 \\
DEV-N & 85.7\,\scriptsize{$\pm$1.2} & 84.6\,\scriptsize{$\pm$5.0} & \textbf{82.8}\,\scriptsize{$\pm$5.0} & 80.9\,\scriptsize{$\pm$9.1} & 82.7\,\scriptsize{$\pm$5.0} & 83.3 & & 76.4\,\scriptsize{$\pm$4.1} & 77.4\,\scriptsize{$\pm$1.4} & 76.9 & & 81.5 & 9.3 \\
\midrule
\textbf{Ours} & \textbf{87.8}\,\scriptsize{$\pm$5.1} & \textbf{85.5}\,\scriptsize{$\pm$4.9} & 82.5\,\scriptsize{$\pm$5.6} & \textbf{85.2}\,\scriptsize{$\pm$5.5} & 86.8\,\scriptsize{$\pm$2.1} & \textbf{85.6} & & \textbf{82.1}\,\scriptsize{$\pm$2.3} & \textbf{81.3}\,\scriptsize{$\pm$0.85} & \textbf{81.7} & & \textbf{84.5} & \textbf{6.4} \\
\bottomrule
\end{tabular}%
}
\end{table}

\noindent\textbf{Ablation Studies.} We ablate three aspects of our method: the aggregation structure underlying the reference construction; the algorithm pool, varying both its size and composition; and the number of checkpoints saved per algorithm.

\begin{table}[h]
\centering
\caption{Ablation on aggregation structure, reported as selected target accuracy (\%). \textit{All checkpoints} builds the reference from every checkpoint with no validator nomination or algorithm structure; \textit{Per-validator} retains validator nomination but ignores algorithm boundaries.  Our two-level design outperforms \textit{All checkpoints} and \textit{Per-validator}.}
\label{tab:abl_aggregation}
\setlength{\tabcolsep}{3pt}
\renewcommand{\arraystretch}{1.1}
\resizebox{\textwidth}{!}{%
\begin{tabular}{l | cccccc | c ccc | c cc}
\toprule
& \multicolumn{6}{c}{\textbf{Brain MRI Scenarios}} & & \multicolumn{3}{c}{\textbf{CXR Scenarios}} & & \multicolumn{2}{c}{\textbf{Overall}} \\
\cmidrule(lr){2-7} \cmidrule(lr){9-11} \cmidrule(lr){13-14}
Aggregation
& A1$\rightarrow$A2 & A1$\rightarrow$A3 & A2$\rightarrow$A1 & A2$\rightarrow$A3 & A1+2$\rightarrow$AIBL & \textbf{Avg.}
& & RSNA$\rightarrow$Child & LDD$\rightarrow$CRD & \textbf{Avg.}
& & All & $\Delta$\textit{Oracle}$\downarrow$ \\
\midrule
\textit{All checkpoints}
& 86.9{\scriptsize$\pm$4.5} & {83.8}{\scriptsize$\pm$4.9} & {81.3}{\scriptsize$\pm$5.3} & {82.0}{\scriptsize$\pm$6.9} & \textbf{86.9}{\scriptsize$\pm$1.9} & {84.1}
& & 81.1{\scriptsize$\pm$2.2} & 79.5{\scriptsize$\pm$0.7} & 80.3
& & {83.0} & {7.8} \\
\textit{Per-validator}
& 86.7{\scriptsize$\pm$4.8} & 80.6{\scriptsize$\pm$7.6} & \textbf{82.6}{\scriptsize$\pm$4.6} & 82.5{\scriptsize$\pm$4.1} & 81.5{\scriptsize$\pm$3.6} & 82.8
& & 78.2{\scriptsize$\pm$5.1} & 79.8{\scriptsize$\pm$2.8} & 79.0
& & 81.7 & 9.1 \\
Two-level (Ours)
& \textbf{87.8}{\scriptsize$\pm$5.1} & \textbf{85.5}{\scriptsize$\pm$4.9} & 82.5{\scriptsize$\pm$5.6} & \textbf{85.2}{\scriptsize$\pm$5.5} & 86.8{\scriptsize$\pm$2.1} & \textbf{85.6}
& & \textbf{82.1}{\scriptsize$\pm$2.3} & \textbf{81.3}{\scriptsize$\pm$0.9} & \textbf{81.7}
& & \textbf{84.5} & \textbf{6.4} \\
\bottomrule
\end{tabular}%
}
\end{table}

\noindent\textit{Aggregation structure.} Table~\ref{tab:abl_aggregation} compares our two-level design with two flat baselines. The first, \textit{all checkpoints}, builds the reference by averaging the predicted class probabilities of every checkpoint, with no validator nomination or algorithm structure. The second, \textit{per-validator}, retains validator nomination but ignores algorithm boundaries by pooling all checkpoints into a single global pool. Our two-level design outperforms \textit{per-validator} and \textit{all checkpoints} varients.

\begin{figure}[h]
    \centering
    \includegraphics[width=1.0\linewidth]{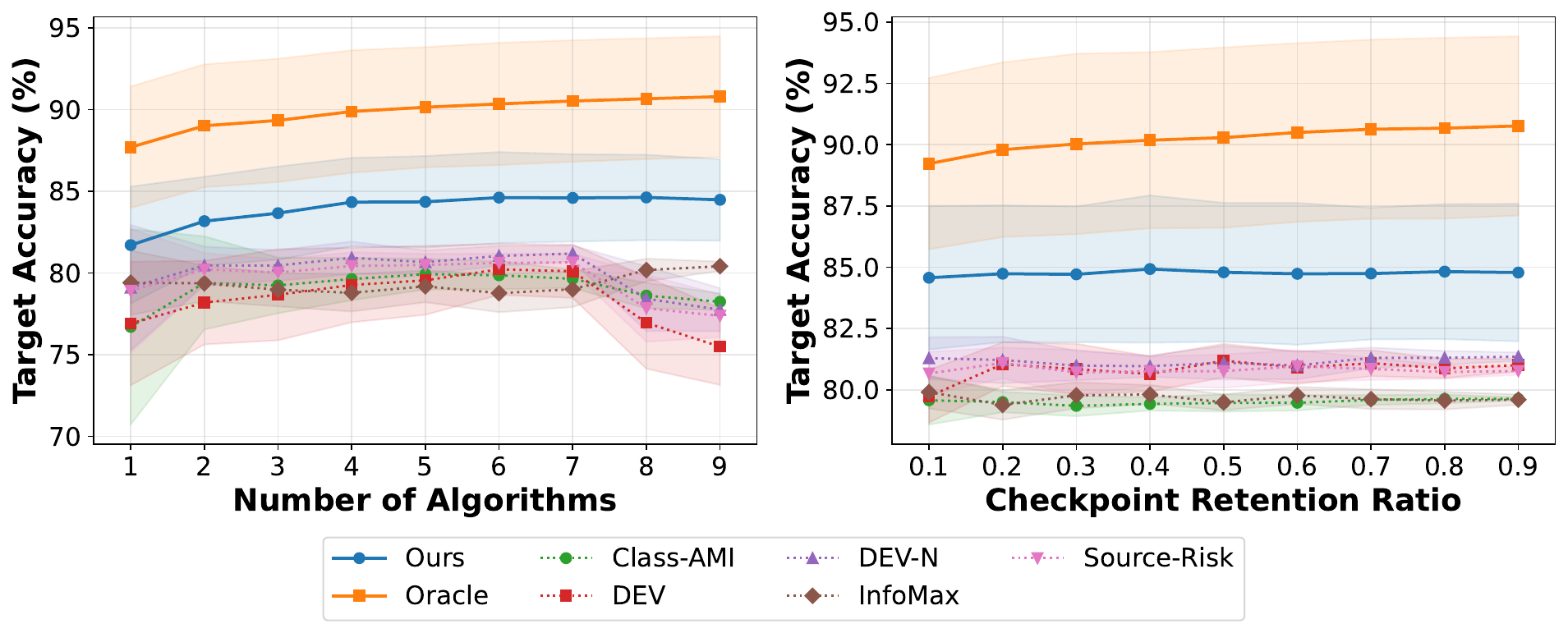}
    \caption{Ablation on the algorithm pool (left) and the fraction of checkpoints retained per algorithm (right), reported as target accuracy (\%) averaged over all UDA scenarios, with shaded regions indicating standard deviation. At each pool size, subsets are sampled randomly, varying both the number and composition of algorithms. Our method maintains a stable gap to \textit{Oracle} and stays above other individual validators across both axes.}
    \label{fig:num_algos_ablation}
\end{figure}

\noindent\textit{Algorithm pool.} Our method builds the agreement reference from checkpoints across multiple algorithms, so a natural question is whether it remains robust under different algorithm pools. To test this, we vary the algorithm pool by randomly sampling several subsets at each size from one to all nine available per scenario (eight general plus one modality-specific), reporting the average over the sampled subsets and all UDA scenarios. Since the subsets are sampled randomly, this varies both the number and the composition of algorithms in the pool. As shown in Figure~\ref{fig:num_algos_ablation}, compared to individual validators, our method maintains a smaller and more stable gap to the \textit{Oracle} across different pools, suggesting that its selection quality holds up well as the pool varies.

\noindent\textit{Number of checkpoints per algorithm.} Our method draws on the checkpoints saved during training, so we also test whether it depends on having many checkpoints per algorithm. We subsample each algorithm's checkpoints at ratios from 0.1 to 0.9, repeating the sampling ten times at each ratio and reporting the average over the sampled draws and all UDA scenarios. As shown in Figure~\ref{fig:num_algos_ablation}, our method keeps a stable gap to the \textit{Oracle} and remains the highest among other validators across different subsample ratios. This suggests that our approach is able to remain effective with relatively few checkpoints per algorithm.

\noindent\textbf{Discussion.} To apply UDA in clinical practice, selecting the best-suited algorithm is particularly difficult: target performance cannot be evaluated directly, and the candidate pool is heterogeneous, spanning not just different hyperparameters but different algorithms. Our two-level agreement reference is designed to address this by drawing on multiple validators within each algorithm and then aggregating across diverse algorithms, with the aim of combining complementary signals rather than relying on any single one. In Table~\ref{tab:abl_aggregation}, we observe that \textit{All checkpoints} outperforms \textit{Per-validator} on average. A possible reason is that, without algorithm boundaries, a validator may nominate checkpoints mainly from algorithms whose training objective is similar to its own criterion. This could reduce the diversity of the reference, whereas using all checkpoints preserves it. Although the reference is built from multiple checkpoints, it is used only to identify a single checkpoint for deployment; the nominated checkpoints do not need to be stored or deployed. Beyond this, the selection quality of our method remains stable across different algorithm pools. We interpret this through two regimes. With few algorithms, the selection task is easier since there are fewer candidates, and the first level of agreement is often enough to identify a good checkpoint. With many algorithms, the task is harder as the candidate space is more heterogeneous, but the larger pool gives the second level of agreement more diversity to draw on, making the reference more reliable. From a clinical deployment perspective, since the best-available checkpoint improves as more algorithms are added, we recommend training multiple UDA algorithms spanning different paradigms when resources allow, giving the method stronger options to select among; when resources are limited, the method still selects reliably from a smaller pool, though deployed performance may be bounded by it.

\noindent\textbf{Limitations and Future Work.} While our method achieves better selection than individual validators, a gap to the \textit{Oracle} remains, most notably on ADNI-1$\rightarrow$ADNI-3. Selecting a single strong checkpoint also requires training potentially multiple algorithms, which is computationally costly; improving the efficiency of this process is an important direction for future work. In addition, our study is limited to binary classification and uses balanced accuracy as the selection target; extending to multi-class classification or segmentation, and to other clinically relevant metrics such as sensitivity, are natural next steps. 

\section{Conclusion}
In this paper, we propose, to our knowledge, the first label-free criterion for joint algorithm and hyperparameter selection in UDA for medical imaging. Our criterion scores each candidate against a two-level agreement reference, built from multiple validators within each algorithm and aggregated across algorithms. The highest-scoring candidate is then selected for deployment. Experimental results on four brain MRI and four CXR datasets across seven clinically relevant transfer scenarios show that our method achieves better selection results than individual validators on average. Our approach also remains effective across different algorithm pools and checkpoint densities. We see this as a step towards practical, label-free algorithm selection for clinical deployment of UDA.

\noindent\textbf{Prospects of Application.} A common scenario could be adapting from labeled data at hospital A to unlabeled data at hospital B, which may differ in scanner, acquisition protocol, or patient population. By applying our method, a practitioner could jointly select the algorithm and hyperparameters from many candidates, obtaining a more reliable deployable model to support physicians in analyzing the data at hospital B.

\begin{credits}
\subsubsection{\ackname} This study was funded by the German Research Foundation DFG (Project: KEMAI, GRK 3012 – 520750254) and by the German Federal Ministry of Research, Technology and Space BMFTR as part of the University Medicine Network 3.0 (Project: RACOON, 01KX2524).

\subsubsection{\discintname}
The authors have no competing interests to declare that are
relevant to the content of this article.
\end{credits}

\bibliographystyle{splncs04}
\bibliography{main.bib}
\end{document}